\documentclass[11pt,a4paper]{article}
\usepackage[a4paper]{geometry}
\usepackage[hyperref]{acl2017}
\usepackage{times}
\usepackage{latexsym}
\usepackage{graphicx}
\usepackage{todonotes}

\usepackage{url}
\usepackage{enumerate}
\usepackage{paralist}
\usepackage{tikz}
\usepackage{tikz-dependency}
\usepackage[small]{caption}
\usepackage{subcaption}
\usepackage{xspace}
\usepackage{amsmath}
\usepackage{amsfonts}
\usepackage{amssymb}
\usepackage{eucal}
\usepackage{multirow}
\usepackage{enumerate}
\usepackage[inline,shortlabels]{enumitem}

\usepackage{resizegather}

\usetikzlibrary{snakes,arrows,shapes}

\aclfinalcopy % Uncomment this line for the final submission
\def\aclpaperid{82} %  Enter the acl Paper ID here

\usepackage{cleveref}
\crefname{section}{\S}{\S\S}
\Crefname{section}{\S}{\S\S}
\crefname{table}{Tab.}{}
\crefname{figure}{Fig.}{}
\crefname{algorithm}{Alg.}{}
\crefname{equation}{Eq.}{}
\crefname{appendix}{App.}{}
\crefformat{section}{\S#2#1#3}  % remove space between section symbol and the number

\usepackage{bm}
\renewcommand{\vec}[1]{{\ensuremath{\boldsymbol{\mathbf{#1}}}}}
\newcommand{\vc}{\vec{c}}
\newcommand{\vw}{\vec{w}}

\newcommand{\va}{\vec{a}}
\newcommand{\vect}[1]{\boldsymbol{#1}} % Uncomment for BOLD vectors.

\usepackage{xspace}
\newcommand{\wtv}[0]{\texttt{word2vec}\xspace}
\newcommand{\qvec}[0]{\textsc{qvec}\xspace}

\title{Frame-Based Continuous Lexical Semantics through Exponential Family Tensor Factorization and Semantic Proto-Roles}

\newcommand{\codeurl}[0]{{\small \url{https://github.com/fmof/tensor-factorization}}\xspace}

\author{%
  Francis Ferraro \and %
  Adam Poliak \and
  Ryan Cotterell \and
  Benjamin Van Durme
  \vspace{0.2em}
  \\
  Center for Language and Speech Processing \\
  Johns Hopkins University \\
  {\tt \{ferraro,azpoliak,ryan.cotterell,vandurme\}@cs.jhu.edu}
}

\date{}

\begin{document}
\maketitle

\begin{abstract}
  We study how different frame annotations complement one another when
  learning continuous lexical semantics.  We learn the
  %continuous semantic
  representations from a tensorized skip-gram
  model that consistently encodes syntactic-semantic content better, %
  with multiple 10\% gains over baselines.
%  semantic and syntactic content.
\end{abstract}

\section{Introduction}
Consider ``Bill'' in \cref{fig:intro-fig}: what is his
involvement with the words ``would try,'' and what does this
involvement \textit{mean}? Word embeddings represent such meaning as
points in a real-valued vector space
\cite{Deerwester90indexingby,mikolov2013efficient}.  These
representations are often learned by exploiting the
frequency that the word cooccurs with contexts, often within a user-defined
window \cite{harris-1954-distributional,turney2010frequency}.
When built from
large-scale sources, like Wikipedia or web crawls, embeddings
capture general characteristics of words and allow for robust
downstream applications \cite{kim:2014:EMNLP2014,das2015gaussian}.

Frame semantics generalize word meanings to that of analyzing
structured and interconnected labeled ``concepts'' and abstractions
\cite{minsky1974,fillmore1976frame,fillmore1982frame}.  These
concepts, or roles, \textit{implicitly} encode expected properties of
that word.  In a frame semantic analysis of \cref{fig:intro-fig}, the
segment ``would try'' \textit{triggers} the \textsc{Attempt} frame,
filling the expected roles \textsc{Agent} and \textsc{Goal} with
``Bill'' and ``the same tactic,'' respectively. %% \footnote{ As seen in
  %% the \textit{segment} ``would try,'' frames may rely on syntax to
  %% account for language-specific constructions. }
While frame semantics provide a structured form for analyzing words
with crisp, categorically-labeled concepts, the encoded properties and
expectations are implicit.  What does it \textit{mean} to fill a
frame's role?

Semantic proto-role (SPR) theory, motivated by
\newcite{dowty1991thematic}'s thematic proto-role theory, offers an
answer to this. SPR replaces categorical roles with
%a very small number of judgments, generally fewer than twenty, about
judgements about multiple underlying properties about
what is likely true of the entity filling the role.  For example, SPR
talks about how likely it is for Bill to be a willing participant in
the \textsc{Attempt}.  The answer to this and other simple judgments
characterize Bill and his involvement. %%  Note that because SPR talks
%% about how \textit{likely} some property is, SPR can represent
%% individual meanings as (very) low dimensional vectors and still
%% maintain a frame-based organization.
Since SPR both captures the likelihood of certain properties and
characterizes roles as groupings of properties, we can view SPR as
representing a type of continuous frame semantics. %

\begin{figure}[t]
  \input{tex/intro_fig}
\end{figure}

We are interested in capturing these SPR-based properties and
expectations within word embeddings. We present a method that learns
frame-enriched embeddings from millions of documents that have been
semantically parsed with multiple different frame analyzers
\cite{ferraro-2014-concretely}. Our method leverages
\newcite{cotterell2017tensor}'s formulation of
\newcite{mikolov2013efficient}'s popular skip-gram model as
exponential family principal component analysis (EPCA) and tensor
factorization. This paper's primary contributions are:
\begin{enumerate*}[(i)]
\item enriching learned word embeddings with multiple, automatically obtained frames from large, disparate corpora; and
\item demonstrating these enriched embeddings better capture SPR-based properties. %
%\item demonstrating and evaluating the improved semantic and syntactic content these embeddings contain. %and
%\item a full generalization of \newcite{cotterell2017tensor}'s tensor factorization method. %
\end{enumerate*}
In so doing, we also generalize \citeauthor{cotterell2017tensor}'s method to arbitrary tensor dimensions. %
This allows us to include an arbitrary amount of semantic information when learning embeddings. %
Our variable-size tensor factorization code is available at \codeurl. %
%We will release our scalable, variable-dimension tensor factorization code. %

%Our experiments demonstrate that including semantic annotations in word embeddings improve supervised and unsupervised tasks.

\section{Frame Semantics and Proto-Roles}
Frame semantics currently used in NLP have a rich history in linguistic literature. %
\newcite{fillmore1976frame}'s frames are based on a word's context and prototypical
concepts that an individual word evokes; they intend to represent the meaning of lexical items by mapping words to real world concepts and shared experiences. %
Frame-based semantics have inspired many semantic annotation schemata and datasets, such as FrameNet \cite{Baker:1998:BFP:980845.980860}, PropBank \cite{palmer2005proposition}, and Verbnet \cite{schuler2005verbnet}, as well as composite resources
\cite{hovy2006ontonotes,palmer2009semlink,banarescu2012abstract}.\footnote{See \newcite{W14-30:2014} for detailed descriptions on frame semantics' contributions to applied NLP tasks.}

\noindent
\textbf{Thematic Roles and Proto Roles} %
These resources map words to their meanings through discrete/categorically labeled frames and roles; sometimes, as in FrameNet, the roles can be very descriptive (e.g., the \textsc{Degree} role for the \textsc{Affirm\_or\_deny} frame), while in other cases, as in PropBank, the roles can be quite general (e.g., \textsc{Arg0}). %
Regardless of the actual schema, the roles are based on thematic roles, which map a predicate's arguments to a semantic representation that makes various semantic distinctions among the arguments \cite{dowty1989semantic}.\footnote{Thematic role theory is rich, and beyond this paper's scope \cite{whitehead1920concept,davidson-semantics-1967,cresswell1973logics,kamp1979events,carlson1984thematic}.
} %
\newcite{dowty1991thematic} claims that thematic role distinctions are not atomic, i.e., they can be deconstructed and analyzed at a lower level. %
Instead of many discrete thematic roles, \newcite{dowty1991thematic} argues for \textit{proto-thematic roles}, e.g.\ \textsc{Proto-Agent} rather than \textsc{Agent}, where distinctions in proto-roles are based on clusterings of logical entailments. %
That is, \textsc{Proto-Agent}s often have certain properties in common, e.g., manipulating other objects or willingly participating in an action; \textsc{Proto-Patient}s are often changed or affected by some action. %
By decomposing the meaning of roles into properties or expectations that can be reasoned about, proto-roles can be seen as including a form of vector representation within structured frame semantics. %

\section{Continuous Lexical Semantics}\label{sec:cls}
Word embeddings represent word meanings as elements of a (real-valued)
vector space
\cite{Deerwester90indexingby}. \newcite{mikolov2013efficient}'s \wtv
methods---skip-gram (SG) and continuous bag of words
(CBOW)---repopularized these methods. %
We focus on SG, which predicts the context $i$ around a word $j$, with learned
representations $\vc_i$ and $\vw_j$, respectively, as
%\vspace{-5pt}
%\begin{equation*}
$
p(\text{context }i \mid \text{word }j)
\propto
\exp\left(\vc_i^\intercal \vw_j\right)
= \exp\left(\vect{1}^\intercal (\vc_i \odot  \vw_j)\right),
$
where $\odot$ is the Hadamard (pointwise) product. %
Traditionally, the context words $i$ are those words within a small window of $j$ and are trained with negative sampling \cite{goldberg2014ns}. %

\subsection{Skip-Gram as Matrix Factorization}
\newcite{levy2014neural}, and subsequently \newcite{DBLP:journals/corr/KeerthiSK15}, showed how vectors learned under SG with the negative sampling are, under certain conditions, the factorization of (shifted) positive pointwise mutual information. %
\newcite{cotterell2017tensor} showed that SG is a form of exponential family PCA that factorizes the matrix of word/context cooccurrence counts (rather than shifted positive PMI values). %
With this interpretation, they generalize SG from matrix to tensor factorization, and provide a theoretical basis for modeling higher-order SG (or additional context, such as morphological features of words) within a word embeddings framework. %%

Specifically, \citeauthor{cotterell2017tensor} recast higher-order SG as maximizing the log-likelihood
%\vspace{-7.5pt}
\begin{align}
\!\!\!\!\sum_{ijk} {\cal X}_{ijk} \log p(\text{context }i \mid \text{word }j, \text{feature }k) \label{eq:hosg-obj} \\
\!\!\!\!= \sum_{ijk} {\cal X}_{ijk} \log \frac{\exp\left(\vect{1}^\intercal (\vc_i \odot  \vw_j \odot \va_k)\right)}{\sum_{i'} \exp\left(\vect{1}^\intercal (\vc_{i'}   \odot \vw_j \odot \va_k)\right)}, \label{eq:skip-tensor}
\end{align}
where ${\cal X}_{ijk}$ is a cooccurrence count 3-tensor of words $j$, surrounding contexts $i$, and features $k$.

%\noindent\textbf{Skip-Gram as $\mathbf{n}$-Tensor Factorization}
\subsection{Skip-Gram as $\mathbf{n}$-Tensor Factorization}\label{sec:ntensor-factorization}
%Here, we generalize \citeauthor{cotterell2017tensor}'s method to arbitrary dimensional tensors. %
When factorizing an $n$-dimensional tensor to include an arbitrary number of $L$ annotations,
we replace \textit{feature} $k$ in Equation~\eqref{eq:hosg-obj} and $\va_k$ in Equation~\eqref{eq:skip-tensor} with each annotation type $l$ and vector  $\vec{\alpha}_l$ included. %
%Correspondingly,
${\cal X}_{i,j,k}$ becomes ${\cal X}_{i,j,l_1, \ldots l_L}$, representing the number of times word $j$ appeared in context $i$ with features $l_1$ through $l_L$. %
%The objective is to maximize
We maximize
%\vspace{-7.5pt}
\begin{gather*}
  \sum_{i,j,l_1,\ldots,l_L} {\cal X}_{i,j,l_1,\ldots,l_L} \log \beta_{i,j,l_1,\ldots,l_L} \label{eq:sem-skip-tensor}  \\
      \beta_{i,j,l_1,\ldots,l_L} \propto
      \exp\left(\vect{1}^\intercal (\vc_i \odot  \vw_j \odot \vec{\alpha}_{l_1} \odot \cdots \odot \vec{\alpha}_{l_L})\right).
      %\exp\left(\vect{1}^\intercal (\vc_i \odot  \vw_j \bigodot \vec{\alpha}_l)\right).
\end{gather*}

\section{Experiments}\label{sec:framecontexts}

%Our end goal is to learn lexical embeddings from millions of newswire and Wikipedia articles automatically annotated with multiple frame semantic parses. %
Our end goal is to use multiple kinds of automatically obtained, ``in-the-wild'' frame semantic parses in order to improve the semantic content---specifically SPR-type information---within learned lexical embeddings. %
We utilize majority portions of the Concretely Annotated New York Times and Wikipedia corpora from \newcite{ferraro-2014-concretely}. %
These have been annotated with three frame semantic parses: FrameNet from \newcite{das-2010-framenet}, and both FrameNet and PropBank from \newcite{wolfe-2016-fnparse}. %
In total, we use nearly five million frame-annotated documents. %
% 1134984 NYT
% 3787500 Wikipedia

\noindent
\textbf{Extracting Counts} %
The baseline extraction we consider is a standard sliding window: for each word $w_j$ seen $\ge T$ times, extract all words $w_i$ two to the left and right of $w_j$. %
These counts, forming a matrix, are then used within standard \wtv. %
We also follow \newcite{cotterell2017tensor} and augment the above with the signed number of tokens separating $w_i$ and $w_j$, e.g., recording that $w_i$ appeared two to the left of $w_j$; these counts form a 3-tensor. %

To turn semantic parses into tensor counts, we first identify relevant information from the parses. %
We consider all parses that are triggered by the target word $w_j$ (seen $\ge T$ times) and that have at least one role filled by some word in the sentence. %
We organize the extraction around roles and what fills them. %
We extract every word $w_r$ that fills all possible triggered frames; each of those frame and role labels; and the distance between filler $w_r$ and trigger $w_j$. %
This process yields a 9-tensor $\mathcal{X}$.\footnote{
  Each record consists of the trigger, a role filler, the number of words between the trigger and filler, and the relevant frame and roles from the three semantic parsers. %
  Being automatically obtained, the parses are overlapping and incomplete; to properly form $\mathcal{X}$, one can implicitly include special $\langle\textrm{NO\_FRAME}\rangle$ and $\langle\textrm{NO\_ROLE}\rangle$ labels as needed. %
} %
Although we \textbf{always} treat the trigger as the ``original'' word (e.g., word $j$, with vector $\vw_j$), later we consider
\begin{inparaenum}[(1)]
\item what to include from $\mathcal{X}$,
\item what to predict (what to treat as the ``context'' word $i$), and
\item what to treat as auxiliary features.
\end{inparaenum}

\noindent
\textbf{Data Discussion} %
The baseline extraction methods result in roughly symmetric target and surrounding word counts. %
This is not the case for the frame extraction. %
Our target words must trigger some semantic parse, so our target words are actually target triggers. %
However, the surrounding context words are those words that fill semantic roles. %
As shown in Table \ref{tab:vocab-sizes}, there are an order-of-magnitude fewer triggers than target words, but up to an order-of-magnitude \textit{more} surrounding words. %

\begin{table}[t]
  \centering
  \resizebox{\columnwidth}{!}{
    \centering
    \begin{tabular}{ccc}
      & \textbf{windowed} & \textbf{frame} \\
      \multirow{2}{*}{\# target words} & 232 & 35.9 (triggers) \\
      & \textit{404} & \textit{45.7} (triggers) \\
      \cline{2-3}
      \# surrounding  & 232 & 531 (role fillers) \\
      words & \textit{404} & \textit{2,305} (role fillers)\\
      %\# FrameNet & --  & 0.8 \\
      %\# PropBank & --  & 5.1\\
    \end{tabular}
  }  
  \caption{%
    Vocabulary sizes, in thousands, extracted from \newcite{ferraro-2014-concretely}'s data with both the standard sliding context window approach (\S\ref{sec:cls}) and the frame-based approach (\S\ref{sec:framecontexts}). Upper numbers (Roman) are for newswire; lower numbers (italics) are Wikipedia. %
    For both corpora, 800 total FrameNet frame types and 5100 PropBank frame types are extracted. %
  }
  \label{tab:vocab-sizes}
\end{table}
\noindent
\textbf{Implementation} %
We generalize \newcite{levy-goldberg:2014:P14-2}'s and \newcite{cotterell2017tensor}'s code to enable any arbitrary dimensional tensor factorization, as described in \S\ref{sec:ntensor-factorization}.
We learn 100-dimensional embeddings for words that appear at least 100 times from 15 negative samples.\footnote{In preliminary experiments, this occurrence threshold did not change the overall conclusions.} %
The implementation is available at \codeurl.

\noindent
\textbf{Metric}
We evaluate our learned (trigger) embeddings \vw{} via \qvec \cite{qvec:emnlp:15}. %
\qvec uses canonical correlation analysis to measure the Pearson correlation between \vw{} and a collection of \textit{oracle} lexical vectors \vec{o}. %
These oracle vectors are derived from a human-annotated resource. %
For \qvec, higher is better: a higher score indicates \vw{} more closely correlates (positively) with \vec{o}. %

\begin{figure*}[t]
  \centering
  \begin{subfigure}{\columnwidth}
    \centering
    \includegraphics[scale=.35]{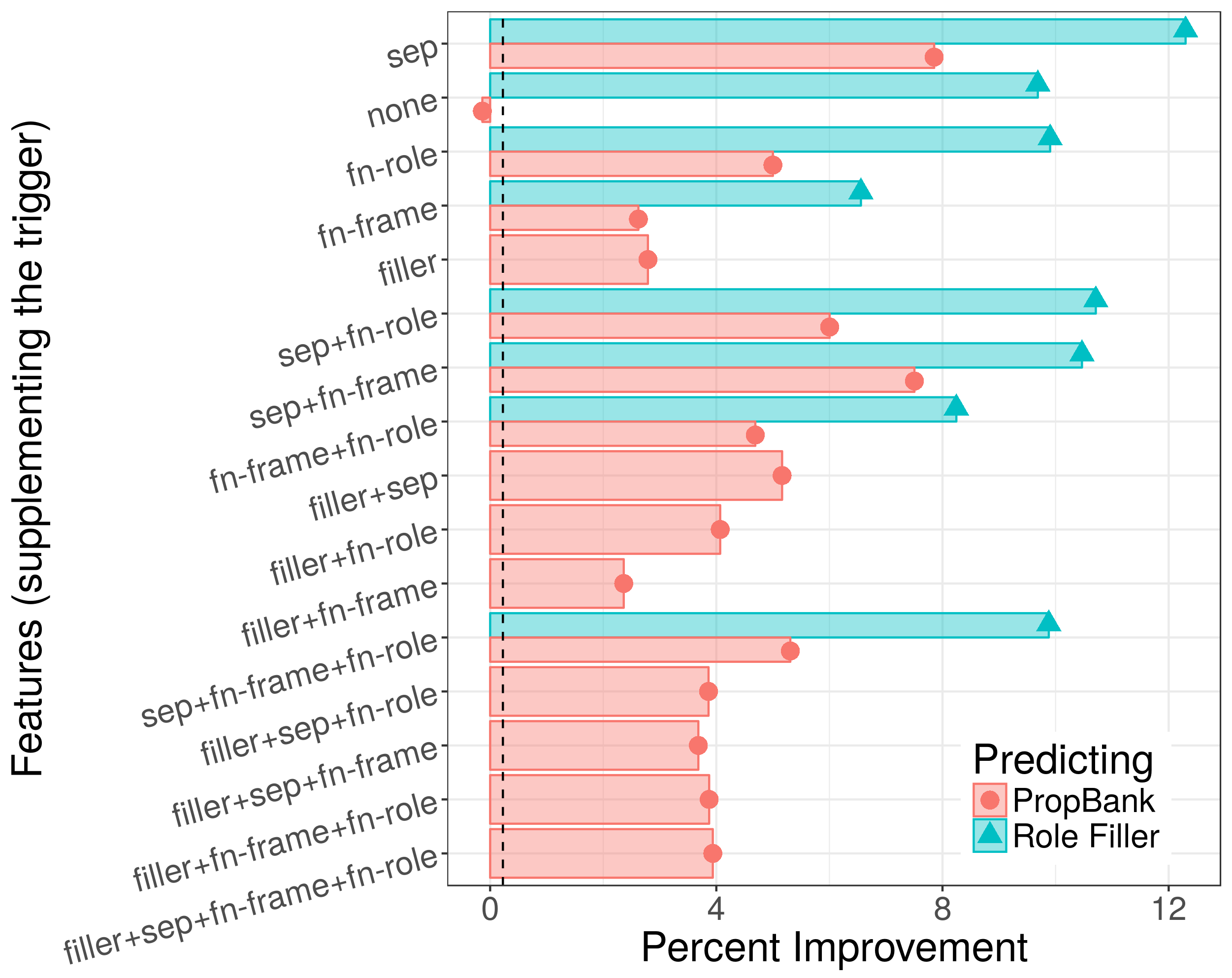}
    \caption{Changes in SPR-\qvec for \textit{Annotated NYT}.}
    \label{fig:anyt-spr-rel-change}
  \end{subfigure}
  ~
  \begin{subfigure}{\columnwidth}
    \centering
    \includegraphics[scale=.35]{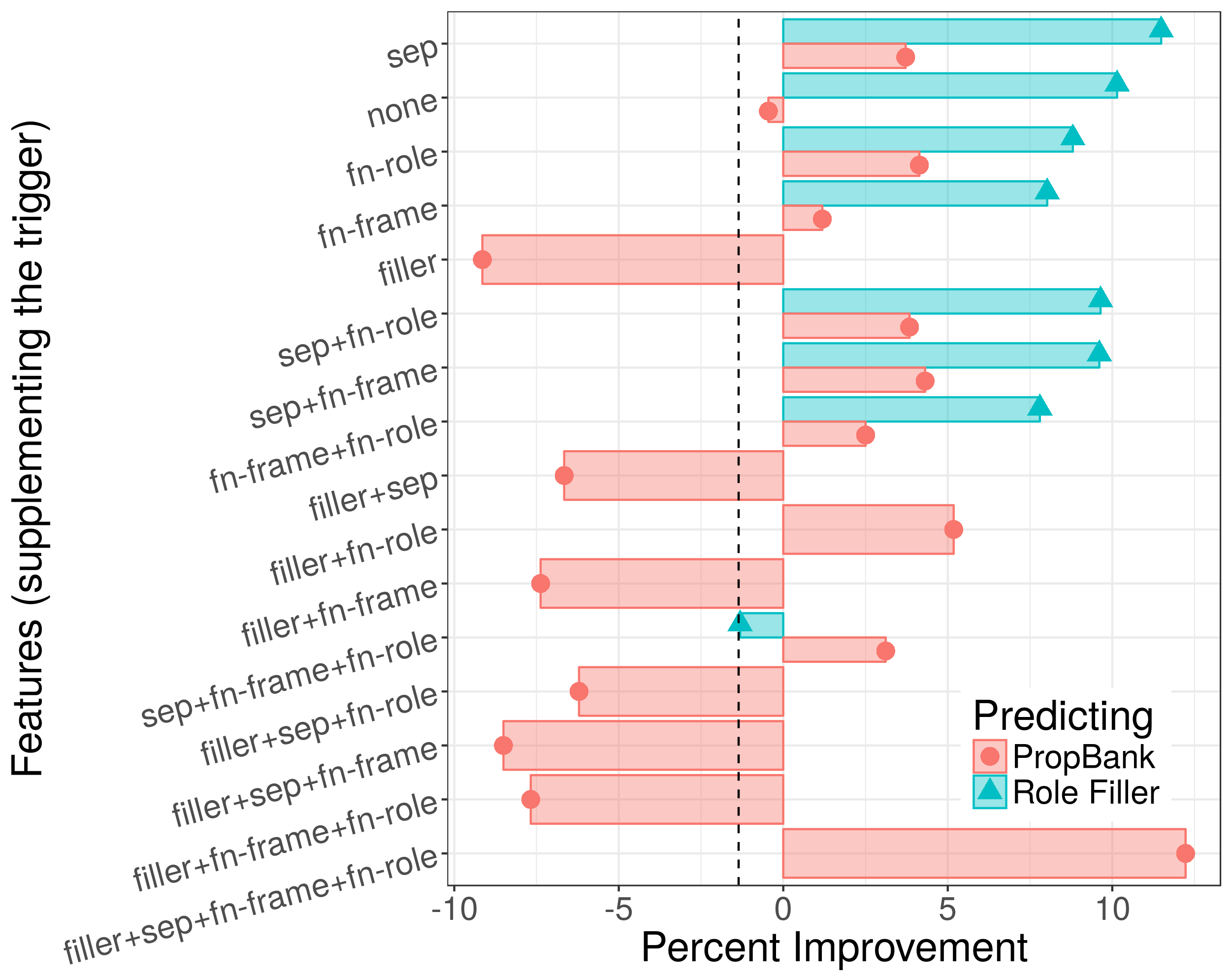}
    \caption{Changes in SPR-\qvec for Wikipedia.}
    \label{fig:wiki-spr-rel-change}
  \end{subfigure}
  \caption{
    Effect of frame-extracted tensor counts on our SPR-\qvec evaulation. %
    Deltas are shown as relative percent changes vs. the \wtv baseline. %
    The dashed line represents the 3-tensor \wtv method of \newcite{cotterell2017tensor}. %
    Each row represents an ablation model: \texttt{sep} means the prediction relies on the token separation distance between the frame and role filler, \texttt{fn-frame} means the prediction uses FrameNet frames, \texttt{fn-role} means the prediction uses FrameNet roles, and \texttt{filler} means the prediction uses the tokens filling the frame role. %
    Read from top to bottom, additional contextual features are denoted with a \texttt{+}. %
    Note when \texttt{filler} is used, we only predict PropBank roles. %
  }
  \label{fig:spr-rel-change}
\end{figure*}

\noindent
\textbf{Evaluating Semantic Content with SPR}
Motivated by \newcite{dowty1991thematic}'s proto-role theory,
\newcite{TACL674}, with a subsequent expansion by \newcite{white-EtAl:2016:EMNLP2016}, annotated thousands of predicate-argument pairs $(v, a)$ with (boolean) applicability and (ordinal) likelihoods of well-motivated semantic properties applying to/being true of $a$.\footnote{
We use the training portion of {\small \url{http://decomp.net/wp-content/uploads/2015/08/UniversalDecompositionalSemantics.tar.gz}}. %
} %
These likelihood judgments, under the SPR framework, are converted from a five-point Likert scale to a 1--5 interval scale. %
Because the predicate-argument pairs were extracted from previously annotated dependency trees, we link each property with the dependency relation joining $v$ and $a$ when forming the oracle vectors; each component of an oracle vector $\vec{o}_v$ is the unity-normalized sum of likelihood judgments for joint property and grammatical relation, using the interval responses when the property is applicable and discarding non-applicable properties, i.e. treating the response as 0. %
Thus, the combined 20 properties of \newcite{TACL674} and \newcite{white-EtAl:2016:EMNLP2016}---together with the four basic grammatical relations \textit{nsubj}, \textit{dobj}, \textit{iobj} and \textit{nsubjpass}---result in 80-dimensional oracle vectors.\footnote{ %
The full cooccurrence among the properties and relations is relatively sparse. %
Nearly two thirds of all non-zero oracle components are comprised of just fourteen properties, and only the \textit{nsubj} and \textit{dobj} relations.} %

\noindent
\textbf{Predict Fillers or Roles?} %
Since SPR judgments are between predicates and arguments, we predict the words filling the roles, and treat all other frame information as auxiliary features. %
SPR annotations were originally based off of (gold-standard) PropBank annotations, so we also train a model to predict PropBank frames and roles, thereby treating role-filling text and all other frame information as auxiliary features. %
In early experiments, we found it beneficial to treat the FrameNet annotations additively and not distinguish one system's output from another. %
Treating the annotations additively serves as a type of collapsing operation. %
Although $\mathcal{X}$ started as a 9-tensor, we only consider up to 6-tensors: trigger, role filler, token separation between the trigger and filler, PropBank frame and role, FrameNet frame, and FrameNet role. %

\noindent
\textbf{Results} %
\cref{fig:spr-rel-change} shows the overall percent change for SPR-\qvec from the filler and role prediction models, on newswire (\cref{fig:anyt-spr-rel-change}) and Wikipedia (\cref{fig:wiki-spr-rel-change}), across different ablation models. %
We indicate additional contextual features being used with a \texttt{+}: \texttt{sep} uses the token separation distance between the frame and role filler, \texttt{fn-frame} uses FrameNet frames, \texttt{fn-role} uses FrameNet roles, \texttt{filler} uses the tokens filling the frame role, and \texttt{none} indicates no additional information is used when predicting. %
The 0 line represents a plain \wtv baseline and the dashed line represents the 3-tensor baseline of \newcite{cotterell2017tensor}. %
Both of these baselines are windowed: they are restricted to a local context and cannot take advantage of frames or any lexical signal that can be derived from frames. %

Overall, we notice that we obtain large improvements from models trained on lexical signals that have been \textit{derived} from frame output (\texttt{sep} and \texttt{none}), even if the model \textit{itself} does not incorporate any frame labels. %
The embeddings that predict the role filling lexical items (the green triangles) correlate higher with SPR oracles than the embeddings that predict PropBank frames and roles (red circles). %
Examining \cref{fig:anyt-spr-rel-change}, we see that both model types outperform both the \wtv and \newcite{cotterell2017tensor} baselines in nearly all model configurations and ablations. %
We see the highest improvement when predicting role fillers given the frame trigger and the number of tokens separating the two (the green triangles in the \texttt{sep} rows). %

Comparing \cref{fig:anyt-spr-rel-change} to \cref{fig:wiki-spr-rel-change}, we see newswire is more amenable to predicting PropBank frames and roles. %
We posit this is a type of out-of-domain error, as the PropBank parser was trained on newswire. %
We also find that newswire is overall more amenable to incorporating limited frame-based features, particularly when predicting PropBank using lexical role fillers as part of the contextual features. %
We hypothesize this is due to the significantly increased vocabulary size of the Wikipedia role fillers (c.f., \cref{tab:vocab-sizes}). %
Note, however, that by using all available schema information when predicting PropBank, we are able to compensate for the increased vocabulary. %

\begin{figure}[t]
  \centering
  \includegraphics[scale=.38]{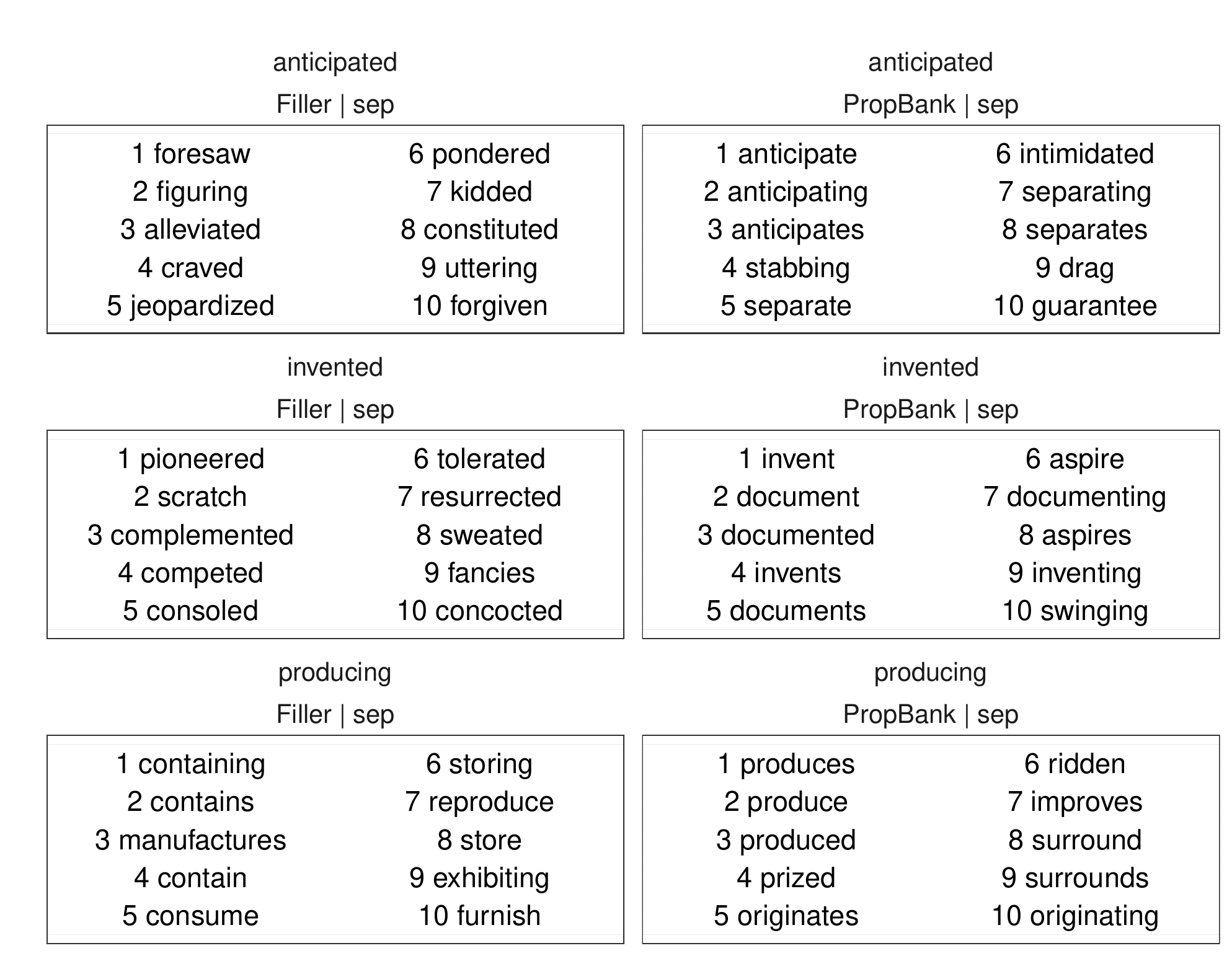}
  \caption{$K$-Nearest Neighbors for three randomly sampled trigger words, from two newswire models.}
  \label{fig:anyt-knn}
\end{figure}

In \cref{fig:anyt-knn} we display the ten nearest neighbors for three randomly sampled trigger words according to two of the highest performing newswire models. %
They each condition on the trigger and the role filler/trigger separation; these correspond to the \texttt{sep} rows of \cref{fig:anyt-spr-rel-change}. %
The left column of \cref{fig:anyt-knn} predicts the role filler, while the right column predicts PropBank annotations. %
We see that while both models learn inflectional relations, this quality is prominent in the model that predicts PropBank information while the model predicting role fillers learns more non-inflectional paraphrases. %

\section{Related Work}
The recent popularity of word embeddings have inspired others to consider leveraging
linguistic annotations and resources to learn embeddings. Both \newcite{cotterell2017tensor}
and \newcite{levy-goldberg:2014:P14-2} incorporate additional syntactic and morphological
information in their word embeddings. %
\newcite{rothe-2015-autoextend}'s use lexical resource entries, such as WordNet synsets, to improve pre-computed word embeddings. %
Through generalized CCA, \newcite{rastogi-2015-mvlsa} incorporate paraphrased FrameNet training data. % to improve
On the applied side, \newcite{wang-2015-petpeeve} used frame embeddings---produced by training \wtv on tweet-derived semantic frame (names)---as additional features in downstream prediction.

\newcite{AAAI1714997} similarly explored the relationship between semantic frames and thematic proto-roles.
They proposed using a Conditional Random Field \cite{Lafferty:2001:CRF:645530.655813} to jointly and conditionally model SPR and SRL\@. %
\newcite{AAAI1714997} demonstrated slight improvements in jointly and conditionally predicting PropBank \cite{bonial-EtAl:2013:LDL2013}'s semantic role
labels and \newcite{TACL674}'s proto-role labels.

\section{Conclusion}
We presented a way to learn embeddings enriched with multiple,
automatically obtained frames from large, disparate corpora. We also
presented a \qvec evaluation for semantic proto-roles. As demonstrated
by our experiments, our extension of \newcite{cotterell2017tensor}'s
tensor factorization enriches word embeddings by including syntactic-semantic information not often captured, resulting in consistently higher SPR-based correlations.
The implementation is available at \codeurl. %
%% semantic and syntactic information often not captured
%% in word embeddings.

\subsection*{Acknowledgments}
This work was supported by Johns Hopkins University, the Human Language Technology Center of Excellence (HLTCOE),
DARPA DEFT, and DARPA LORELEI\@. %
We would also like to thank three anonymous reviewers for their feedback. %
The views and conclusions
contained in this publication are those of the authors and should not
be interpreted as representing official policies or endorsements of
DARPA or the U.S. Government.

\bibliography{sem_hsog}

\begin{thebibliography}{}
\expandafter\ifx\csname natexlab\endcsname\relax\def\natexlab#1{#1}\fi

\bibitem[{Baker et~al.(1998)Baker, Fillmore, and
  Lowe}]{Baker:1998:BFP:980845.980860}
Collin~F. Baker, Charles~J. Fillmore, and John~B. Lowe. 1998.
\newblock \href{https://doi.org/10.3115/980845.980860}{The berkeley framenet
  project}.
\newblock In {\em Proceedings of the 36th Annual Meeting of the Association for
  Computational Linguistics and 17th International Conference on Computational
  Linguistics - Volume 1\/}. Association for Computational Linguistics,
  Stroudsburg, PA, USA, ACL '98, pages 86--90.
\newblock
  \href{https://doi.org/10.3115/980845.980860}{https://doi.org/10.3115/980845.980860}.

\bibitem[{Banarescu et~al.(2012)Banarescu, Bonial, Cai, Georgescu, Griffitt,
  Hermjakob, Knight, Koehn, Palmer, and Schneider}]{banarescu2012abstract}
Laura Banarescu, Claire Bonial, Shu Cai, Madalina Georgescu, Kira Griffitt, Ulf
  Hermjakob, Kevin Knight, Philipp Koehn, Martha Palmer, and Nathan Schneider.
  2012.
\newblock Abstract meaning representation (amr) 1.0 specification.
\newblock In {\em Parsing on Freebase from Question-Answer Pairs. In
  Proceedings of the 2013 Conference on Empirical Methods in Natural Language
  Processing. Seattle: ACL\/}. pages 1533--1544.

\bibitem[{Bonial et~al.(2013)Bonial, Stowe, and
  Palmer}]{bonial-EtAl:2013:LDL2013}
Claire Bonial, Kevin Stowe, and Martha Palmer. 2013.
\newblock \href{http://www.aclweb.org/anthology/W13-5503}{Renewing and revising
  semlink}.
\newblock In {\em Proceedings of the 2nd Workshop on Linked Data in Linguistics
  (LDL-2013): Representing and linking lexicons, terminologies and other
  language data\/}. Association for Computational Linguistics, Pisa, Italy,
  pages 9 -- 17.
\newblock
  \href{http://www.aclweb.org/anthology/W13-5503}{http://www.aclweb.org/anthology/W13-5503}.

\bibitem[{Carlson(1984)}]{carlson1984thematic}
Greg~N Carlson. 1984.
\newblock Thematic roles and their role in semantic interpretation.
\newblock {\em Linguistics\/} 22(3):259--280.

\bibitem[{Cotterell et~al.(2017)Cotterell, Poliak, Van~Durme, and
  Eisner}]{cotterell2017tensor}
Ryan Cotterell, Adam Poliak, Benjamin Van~Durme, and Jason Eisner. 2017.
\newblock Explaining and generalizing skip-gram through exponential family
  principal component analysis.
\newblock In {\em Proceedings of the 15th Conference of the European Chapter of
  the Association for Computational Linguistics\/}. Valencia, Spain.

\bibitem[{Cresswell(1973)}]{cresswell1973logics}
Maxwell~John Cresswell. 1973.
\newblock {\em Logics and languages\/}.
\newblock London: Methuen [Distributed in the U.S.A. By Harper \& Row].

\bibitem[{Das et~al.(2010)Das, Schneider, Chen, and Smith}]{das-2010-framenet}
Dipanjan Das, Nathan Schneider, Desai Chen, and Noah~A Smith. 2010.
\newblock Probabilistic frame-semantic parsing.
\newblock In {\em Human language technologies: The 2010 annual conference of
  the North American chapter of the association for computational
  linguistics\/}. Association for Computational Linguistics, pages 948--956.

\bibitem[{Das et~al.(2015)Das, Zaheer, and Dyer}]{das2015gaussian}
Rajarshi Das, Manzil Zaheer, and Chris Dyer. 2015.
\newblock \href{http://www.aclweb.org/anthology/P15-1077}{Gaussian lda for
  topic models with word embeddings}.
\newblock In {\em Proceedings of the 53rd Annual Meeting of the Association for
  Computational Linguistics and the 7th International Joint Conference on
  Natural Language Processing (Volume 1: Long Papers)\/}. Association for
  Computational Linguistics, Beijing, China, pages 795--804.
\newblock
  \href{http://www.aclweb.org/anthology/P15-1077}{http://www.aclweb.org/anthology/P15-1077}.

\bibitem[{Davidson(1967)}]{davidson-semantics-1967}
Donald Davidson. 1967.
\newblock The logical form of action sentences.
\newblock In Nicholas Rescher, editor, {\em The Logic of Decision and
  Action\/}, University of Pittsburgh Press.

\bibitem[{Deerwester et~al.(1990)Deerwester, Dumais, Furnas, Landauer, and
  Harshman}]{Deerwester90indexingby}
Scott Deerwester, Susan~T. Dumais, George~W. Furnas, Thomas~K. Landauer, and
  Richard Harshman. 1990.
\newblock Indexing by latent semantic analysis.
\newblock {\em JOURNAL OF THE AMERICAN SOCIETY FOR INFORMATION SCIENCE\/}
  41(6):391--407.

\bibitem[{Dowty(1991)}]{dowty1991thematic}
David Dowty. 1991.
\newblock Thematic proto-roles and argument selection.
\newblock {\em Language\/} 67(3):547--619.

\bibitem[{Dowty(1989)}]{dowty1989semantic}
David~R Dowty. 1989.
\newblock On the semantic content of the notion of ‘thematic role’.
\newblock In {\em Properties, types and meaning\/}, Springer, pages 69--129.

\bibitem[{Ferraro et~al.(2014)Ferraro, Thomas, Gormley, Wolfe, Harman, and {Van
  Durme}}]{ferraro-2014-concretely}
Francis Ferraro, Max Thomas, Matthew~R. Gormley, Travis Wolfe, Craig Harman,
  and Benjamin {Van Durme}. 2014.
\newblock {C}oncretely {A}nnotated {C}orpora.
\newblock In {\em 4th Workshop on Automated Knowledge Base Construction
  (AKBC)\/}.

\bibitem[{Fillmore(1982)}]{fillmore1982frame}
Charles Fillmore. 1982.
\newblock Frame semantics.
\newblock {\em Linguistics in the morning calm\/} pages 111--137.

\bibitem[{Fillmore(1976)}]{fillmore1976frame}
Charles~J Fillmore. 1976.
\newblock Frame semantics and the nature of language*.
\newblock {\em Annals of the New York Academy of Sciences\/} 280(1):20--32.

\bibitem[{Goldberg and Levy(2014)}]{goldberg2014ns}
Yoav Goldberg and Omer Levy. 2014.
\newblock word2vec explained: {D}eriving {M}ikolov et al.'s negative-sampling
  word-embedding method.
\newblock {\em arXiv preprint arXiv:1402.3722\/} .

\bibitem[{Harris(1954)}]{harris-1954-distributional}
Zellig~S Harris. 1954.
\newblock Distributional structure.
\newblock {\em Word\/} 10(2-3):146--162.

\bibitem[{Hovy et~al.(2006)Hovy, Marcus, Palmer, Ramshaw, and
  Weischedel}]{hovy2006ontonotes}
Eduard Hovy, Mitchell Marcus, Martha Palmer, Lance Ramshaw, and Ralph
  Weischedel. 2006.
\newblock Ontonotes: the 90\% solution.
\newblock In {\em Proceedings of the human language technology conference of
  the NAACL, Companion Volume: Short Papers\/}. Association for Computational
  Linguistics, pages 57--60.

\bibitem[{Kamp(1979)}]{kamp1979events}
Hans Kamp. 1979.
\newblock Events, instants and temporal reference.
\newblock In {\em Semantics from different points of view\/}, Springer, pages
  376--418.

\bibitem[{Keerthi et~al.(2015)Keerthi, Schnabel, and
  Khanna}]{DBLP:journals/corr/KeerthiSK15}
S.~Sathiya Keerthi, Tobias Schnabel, and Rajiv Khanna. 2015.
\newblock Towards a better understanding of predict and count models.
\newblock {\em arXiv preprint arXiv:1511.0204\/} .

\bibitem[{Kim(2014)}]{kim:2014:EMNLP2014}
Yoon Kim. 2014.
\newblock \href{http://www.aclweb.org/anthology/D14-1181}{Convolutional neural
  networks for sentence classification}.
\newblock In {\em Proceedings of the 2014 Conference on Empirical Methods in
  Natural Language Processing (EMNLP)\/}. Association for Computational
  Linguistics, Doha, Qatar, pages 1746--1751.
\newblock
  \href{http://www.aclweb.org/anthology/D14-1181}{http://www.aclweb.org/anthology/D14-1181}.

\bibitem[{Lafferty et~al.(2001)Lafferty, McCallum, and
  Pereira}]{Lafferty:2001:CRF:645530.655813}
John~D. Lafferty, Andrew McCallum, and Fernando C.~N. Pereira. 2001.
\newblock \href{http://dl.acm.org/citation.cfm?id=645530.655813}{Conditional
  random fields: Probabilistic models for segmenting and labeling sequence
  data}.
\newblock In {\em Proceedings of the Eighteenth International Conference on
  Machine Learning\/}. Morgan Kaufmann Publishers Inc., San Francisco, CA, USA,
  ICML '01, pages 282--289.
\newblock
  \href{http://dl.acm.org/citation.cfm?id=645530.655813}{http://dl.acm.org/citation.cfm?id=645530.655813}.

\bibitem[{Levy and Goldberg(2014{\natexlab{a}})}]{levy-goldberg:2014:P14-2}
Omer Levy and Yoav Goldberg. 2014{\natexlab{a}}.
\newblock \href{http://www.aclweb.org/anthology/P14-2050}{Dependency-based word
  embeddings}.
\newblock In {\em Proceedings of the 52nd Annual Meeting of the Association for
  Computational Linguistics (Volume 2: Short Papers)\/}. Association for
  Computational Linguistics, Baltimore, Maryland, pages 302--308.
\newblock
  \href{http://www.aclweb.org/anthology/P14-2050}{http://www.aclweb.org/anthology/P14-2050}.

\bibitem[{Levy and Goldberg(2014{\natexlab{b}})}]{levy2014neural}
Omer Levy and Yoav Goldberg. 2014{\natexlab{b}}.
\newblock Neural word embedding as implicit matrix factorization.
\newblock In {\em Advances in neural information processing systems\/}. pages
  2177--2185.

\bibitem[{Mikolov et~al.(2013)Mikolov, Chen, Corrado, and
  Dean}]{mikolov2013efficient}
Tomas Mikolov, Kai Chen, Greg Corrado, and Jeffrey Dean. 2013.
\newblock Efficient estimation of word representations in vector space.
\newblock {\em arXiv preprint arXiv:1301.3781\/} .

\bibitem[{Minsky(1974)}]{minsky1974}
Marvin Minsky. 1974.
\newblock A framework for representing knowledge. {MIT-AI} {L}aboratory {M}emo
  306.

\bibitem[{Palmer(2009)}]{palmer2009semlink}
Martha Palmer. 2009.
\newblock Semlink: Linking propbank, verbnet and framenet.
\newblock In {\em Proceedings of the Generative Lexicon Conference\/}.
  GenLex-09, 2009 Pisa, Italy, pages 9--15.

\bibitem[{Palmer et~al.(2005)Palmer, Gildea, and
  Kingsbury}]{palmer2005proposition}
Martha Palmer, Daniel Gildea, and Paul Kingsbury. 2005.
\newblock The proposition bank: An annotated corpus of semantic roles.
\newblock {\em Computational linguistics\/} 31(1):71--106.

\bibitem[{Petruck and {de Melo}(2014)}]{W14-30:2014}
Miriam R.~L. Petruck and Gerard {de Melo}, editors. 2014.
\newblock {\em Proceedings of Frame Semantics in NLP: A Workshop in Honor of
  Chuck Fillmore (1929-2014)\/}.
\newblock Association for Computational Linguistics, Baltimore, MD, USA.
\newblock
  \href{http://www.aclweb.org/anthology/W14-30}{http://www.aclweb.org/anthology/W14-30}.

\bibitem[{Rastogi et~al.(2015)Rastogi, Van~Durme, and
  Arora}]{rastogi-2015-mvlsa}
Pushpendre Rastogi, Benjamin Van~Durme, and Raman Arora. 2015.
\newblock \href{http://www.aclweb.org/anthology/N15-1058}{Multiview {LSA}:
  {R}epresentation {L}earning via {G}eneralized {CCA}}.
\newblock In {\em Proceedings of the 2015 Conference of the North American
  Chapter of the Association for Computational Linguistics: Human Language
  Technologies\/}. Association for Computational Linguistics, Denver, Colorado,
  pages 556--566.
\newblock
  \href{http://www.aclweb.org/anthology/N15-1058}{http://www.aclweb.org/anthology/N15-1058}.

\bibitem[{Reisinger et~al.(2015)Reisinger, Rudinger, Ferraro, Harman, Rawlins,
  and Durme}]{TACL674}
Drew Reisinger, Rachel Rudinger, Francis Ferraro, Craig Harman, Kyle Rawlins,
  and Benjamin~Van Durme. 2015.
\newblock Semantic proto-roles.
\newblock {\em Transactions of the Association for Computational Linguistics
  (TACL)\/} 3:475--488.

\bibitem[{Rothe and Sch\"{u}tze(2015)}]{rothe-2015-autoextend}
Sascha Rothe and Hinrich Sch\"{u}tze. 2015.
\newblock \href{http://www.aclweb.org/anthology/P15-1173}{Autoextend: Extending
  word embeddings to embeddings for synsets and lexemes}.
\newblock In {\em Proceedings of the 53rd Annual Meeting of the Association for
  Computational Linguistics and the 7th International Joint Conference on
  Natural Language Processing (Volume 1: Long Papers)\/}. Association for
  Computational Linguistics, Beijing, China, pages 1793--1803.
\newblock
  \href{http://www.aclweb.org/anthology/P15-1173}{http://www.aclweb.org/anthology/P15-1173}.

\bibitem[{Schuler(2005)}]{schuler2005verbnet}
Karin~Kipper Schuler. 2005.
\newblock Verbnet: A broad-coverage, comprehensive verb lexicon .

\bibitem[{Teichert et~al.(2017)Teichert, Poliak, Durme, and
  Gormley}]{AAAI1714997}
Adam Teichert, Adam Poliak, Benjamin~Van Durme, and Matthew Gormley. 2017.
\newblock Semantic proto-role labeling.
\newblock In {\em AAAI Conference on Artificial Intelligence\/}.

\bibitem[{Tsvetkov et~al.(2015)Tsvetkov, Faruqui, Ling, Lample, and
  Dyer}]{qvec:emnlp:15}
Yulia Tsvetkov, Manaal Faruqui, Wang Ling, Guillaume Lample, and Chris Dyer.
  2015.
\newblock \href{http://aclweb.org/anthology/D15-1243}{Evaluation of word vector
  representations by subspace alignment}.
\newblock In {\em Proceedings of the 2015 Conference on Empirical Methods in
  Natural Language Processing\/}. Association for Computational Linguistics,
  Lisbon, Portugal, pages 2049--2054.
\newblock
  \href{http://aclweb.org/anthology/D15-1243}{http://aclweb.org/anthology/D15-1243}.

\bibitem[{Turney and Pantel(2010)}]{turney2010frequency}
Peter~D Turney and Patrick Pantel. 2010.
\newblock From frequency to meaning: Vector space models of semantics.
\newblock {\em Journal of artificial intelligence research\/} 37:141--188.

\bibitem[{Wang and Yang(2015)}]{wang-2015-petpeeve}
William~Yang Wang and Diyi Yang. 2015.
\newblock \href{http://aclweb.org/anthology/D15-1306}{That's so annoying!!!: A
  lexical and frame-semantic embedding based data augmentation approach to
  automatic categorization of annoying behaviors using \#petpeeve tweets}.
\newblock In {\em Proceedings of the 2015 Conference on Empirical Methods in
  Natural Language Processing\/}. Association for Computational Linguistics,
  Lisbon, Portugal, pages 2557--2563.
\newblock
  \href{http://aclweb.org/anthology/D15-1306}{http://aclweb.org/anthology/D15-1306}.

\bibitem[{White et~al.(2016)White, Reisinger, Sakaguchi, Vieira, Zhang,
  Rudinger, Rawlins, and Van~Durme}]{white-EtAl:2016:EMNLP2016}
Aaron~Steven White, Drew Reisinger, Keisuke Sakaguchi, Tim Vieira, Sheng Zhang,
  Rachel Rudinger, Kyle Rawlins, and Benjamin Van~Durme. 2016.
\newblock \href{https://aclweb.org/anthology/D16-1177}{Universal
  decompositional semantics on universal dependencies}.
\newblock In {\em Proceedings of the 2016 Conference on Empirical Methods in
  Natural Language Processing\/}. Association for Computational Linguistics,
  Austin, Texas, pages 1713--1723.
\newblock
  \href{https://aclweb.org/anthology/D16-1177}{https://aclweb.org/anthology/D16-1177}.

\bibitem[{Whitehead(1920)}]{whitehead1920concept}
Alfred~North Whitehead. 1920.
\newblock {\em The concept of nature: the Tarner lectures delivered in Trinity
  College, November 1919\/}.
\newblock Kessinger Publishing.

\bibitem[{Wolfe et~al.(2016)Wolfe, Dredze, and Van~Durme}]{wolfe-2016-fnparse}
Travis Wolfe, Mark Dredze, and Benjamin Van~Durme. 2016.
\newblock \href{http://aclweb.org/anthology/W16-5905}{A study of imitation
  learning methods for semantic role labeling}.
\newblock In {\em Proceedings of the Workshop on Structured Prediction for
  NLP\/}. Association for Computational Linguistics, Austin, TX, pages 44--53.
\newblock
  \href{http://aclweb.org/anthology/W16-5905}{http://aclweb.org/anthology/W16-5905}.

\end{thebibliography}
\bibliographystyle{acl_natbib}
\appendix

\end{document}